# Traffic and Mobility Optimization Using AI: Comparative Study between Dubai and Riyadh

KanwalAalijah
School of Electrical Engineering and Computer Sciences (SEECS) National University of Sciences and Technology (NUST)
Islamabad, Pakistan
ksair.mscs18seecs@seecs.edu.pk

*Abstract*—Urban planning plays a very important role in development modern cities. It effects the economic growth, quality of life, and environmental sustainability. Modern cities face challenges in managing traffic congestion. These challenges arise to due to rapid urbanization. In this study we will explore how AI can be used to understand the traffic and mobility related issues and its effects on the resident's sentiment. The approach combines real-time traffic data with geo-located sentiment analysis, offering a comprehensive and dynamic approach to urban mobility planning. AI models and exploratory data analysis was used to predict traffic congestion patterns, analyze commuter behaviors, and identify congestion hotspots and dissatisfaction zones. The findings offer actionable recommendations for optimizing traffic flow, enhancing commuter experiences, and addressing city-specific mobility challenges in the Middle East and beyond.

## I. Introduction

Urban mobility is the heart of sustainable city development, it directly influences the economic growth, quality of life, and environmental health of the residents. As population increases in metropolitan hubs like Dubai and Riyadh continue to rise, the load on the existing transport infrastructures intensifies. Both the cities, are recognized for their rapid urbanization and strategic importance in the Middle East in the recent years. They are currently facing unique challenges in managing traffic congestion and mobility. Addressing these challenges is imperative for maintaining urban functionality and achieving the broader visions of development—Dubai's Smart City 2021 initiative and Saudi Arabia's Vision 2030 initiative [1], [2].

AI can be used for traffic and mobility management. It can help the urban planners to create solutions and help them with faster decision-making. AI can be used to efficiently manage traffic flow with applications ranging from real-time optimization of traffic to dynamic routing for a better traffic management. This approach can remove the traffic congestion in the city and may also reduce the emissions thereby improving the experience of the daily commuters [3], [4]. Application of AI in urbanization can vary from city to city and it is unique for each city [5].

Dubai as a city sets itself apart from Riyadh. Dubai has expanded vertically mostly and whereas Riyadh has horizontally scaling. Dubai has better public transport system where as Riyadh relies upon car-based mobility [4], [6]. This aspect presents a very unique chance to evaluate the efficiency of the AI based models in improving traffic mobility in both the urban cities. Dubai focuses on technological infrastructures, including autonomous vehicles and IoT-enabled systems whereas Riyadh is currently focusing on developing better public transport system [7], [8]. However, both the cities can benefit from big data decision making methods [9]. Big data management techniques [10] are particularly useful when the data belonging to different sectors overlap to understand complex urbanization in cities.

This study aims to understand the traffic hotpot of both the cities and to overlap the sentiment of city resident's to understand how AI can be used to understand and optimize mobility in both the cities.

We leveraged the mobility data from TomTom's comprehensive traffic dataset. In order to understand the sentiment of the residents we crawled the data from popular social media websites such as Tik-tok, X, Facebook, Google reviews etc. [2], [6]. In this research to we are going to analyze traffic congestion patterns, predict commuter behaviors, and perform predictive analysis of the traffic and residents sentiments. The findings are expected to provide actionable recommendations for enhancing urban mobility, with broader implications for other densely populated cities in the Middle East and beyond [1], [4].

## II. Literature Review

Artificial Intelligence can be used for improving urban traffic and mobility systems. Persistent issues of traffic congestion, delays, and environmental sustainability can be understood by using AI. Real-time data and social sentiment analysis allow for new insights to be applied to urban planning which could help address some previously unmet challenges.

AI-based solutions have greatly advanced the traffic forecasting, surveillance, and control of city traffic. For example, neural networks and decision trees work for algorithms based models for traffic prediction, congestion pattern recognition and effective routing [3][4]. These models are also more effective as they use historical data to create plans of action that motivate interventions such as traffic light adjustment and incident management [11].

AI is embedded in smart infrastructure in cities like Dubai. Self-driving cars and IoT-enabled gadgets enhance traffic management and ensure passenger safety [7]. AI models implemented in Dubai have improved travel times and promoted sustainable modes of transport. Riyadh is expanding horizontally, hence there is a lot of traffic build up currently in the city. However, the city is exploring different ways in which AI can be implemented in assisting and optimizing the public transport systems and reducing the congestion hotspots [8], [12].

The incorporation of social media data on traffic and mobility management has alleviated some dimensions for AI models. Geo-tagged posts description, google reviews, and comments of residents on the social media websites can provide valuable insights into real-time public sentiment regarding traffic conditions and urban infrastructure [4]. By applying natural language processing (NLP) techniques, urban planners can map commuter sentiment to specific locations and can identify the pain points such as heavily congested roads or inefficient public transport.

Commuter preferences and activities can be better understood with the help of sentiment analysis tools. For instance, posts with traffic-related keywords, including tweets, can give insights into the resident's opinion on services such as congestion, delays, and transport provisions [5], [13]. While analyzing data during public emergency of COVID-19, shift of public sentiment and emotion toward different transport options was evident and in those social media posts there was a concern for safety [13]. These social medias posts gave an insight on thoughts of residents and their sentiments towards various situations in the city. Dubai and Riyadh governments and urban planner could benefit if resident sentiment data is overlapped with traffic data. This approach could bring the planners closer to solving the issues in citizens-oriented mobility planning.

There are many challenges, when it comes to integrating the social media data with AI. In social media data the geo-location, demographics are not usually tagged [6]. AI models are predicting these demographics and geo-location and their accuracy greatly depends upon the accuracy of the model itself. Additionally, ensuring data privacy and addressing ethical concerns is also very critical when using user-generated for analysis.

Traffic mobility and congestion problems can be studied from a different angle of sentiment analysis. Cities can adopt multi-modal approaches. Tradiational traffic metrics and real-time public sentiment can be combined to develop adaptive and modern urban mobility and planning workflows. Recent trials of AI-based adaptive traffic lights, prioritizing cyclists and reducing congestion, further underscore the transformative potential of AI in creating efficient and sustainable transport systems [11]. Such approaches are particularly relevant for addressing the unique mobility challenges of cities like Dubai and Riyadh, which are striving to balance technological advancements with rapid urban growth.

In summary, the overlap of traffic mobility data and social media analytics presents a promising scenario for enhancing urban planning and management. By using these technologies, cities can develop more responsive and efficient transportation systems.

## III. METHODOLOGY

The Methodology section will outline the steps and tools and AI models used to analyze traffic and traffic mobility in Dubai and Riyadh.

### A. Data Collection

- **Traffic Data:** In this study we have utilized TomTom dataset [17]. The dataset provides granular traffic data for Dubai and Riyadh. The dataset includes Traffic index, Total length and count of traffic jams and delay times and live travel times per 10 kilometers.

- **Social Media Data:** We collected geo-tagged tweets and posts from various social media websites such as X, Facebook, Instagram, Tiktok, google reviews etc. The posts and reviews were collected on the basis of keywords and hashtags such as "traffic", "congestion in Dubai" etc. [14].

### B. Data Preprocessing

- Traffic data was normalized for the metrics to ensure comparability across both cities.

- Social media data was preprocessed that involved several steps such as removing noise (URLs, emojis), normalizing text (standardizing Arabic script), tokenizing text, filtering stop words, and applying stemming and lemmatization to handle Arabic morphology. To capture the context of the Arabic morphology and phrases (bigrams and trigrams) [15], Additionally, n-grams were generated to capture context and common phrases within the short social media posts [15]. After that sentiment analysis was performed on the social media dataset [14], [15].

### C. Exploratory Data Analysis (EDA)

We performed exploratory data analysis (EDA) to understand various traffic and mobility trends in Dubai and Riyadh.

**Traffic Patterns and Metrics:** Various traffic patterns and metrics were uncovered in this analysis. Metrics such as Traffic index, Traffic jams in kilometers, and delay in minutes were visualized in order to understand the peak congestion times and recurring patterns. Various heat maps were generated to display spatial distributions of traffic jams and average congestion levels and delays for the both cities. Live data as well as the historical data for both the cities was also compared for all these metrics.

**Correlation Analysis:** We also performed statistical correlations between congestion indicators (e.g., Jams Count) and travel delays. This approach helped us to identify which factors contributed most to prolonged delays in each city. We correlated social media sentiment data with traffic metrics to explore the relationship between residents' perceptions and real-world traffic conditions, pinpointing areas of dissatisfaction [4], [14]. Correlation analysis was also performed on Travel time, congestion level analysis on travel and speed analysis on Travel time.

**Sentiment Trends:** Sentiment analysis was performed on the posts collected on social media posts [14],[15]. Posts were classified in to sentiment bins; positive, negative and neutral. This was done to understand the alignment of resident sentiment with live traffic conditions [4]. The sentiment trends were then compared against traffic data to identify overlaps between public sentiment and high-traffic areas [15].

### D. AI Modeling Approaches

AI predictive models were then used to predict the traffic patterns as well as the sentiment trends were predicted.

- Long Short-Term Memory (LSTM) networks were used to predict traffic congestion indices and delays. The historical data was used in training the model and then predicting the traffic index, delay and sentiment [16].

- Regression Models were used to predict delay times using the feature of traffic jam count, live traffic index and historical travel time per 10 KM [16].
- Geo-tagged sentiment data was overlaid with traffic metrics to create sentiment heatmaps. These visualizations highlighted areas with high congestion and commuter dissatisfaction, providing actionable insights for urban planners [14] [15].

*E. Evaluation Metrics*

Following metrics were used to evaluate the performance of traffic prediction models, analytical models, sentiment classification system:

- To evaluate the **traffic prediction** models we used metrics such as **Root Mean Squared Error**, **Mean Absolute Error**, and **R-Squared** score were used. RMSE measures the average magnitude of prediction errors, MAE represents average error in predictions, R-Squared score was used to evaluate how well the model explained the variance in traffic metrics.
- Precision, recall and F1 score [15] were used to measure the proportion of correctly identified posts by the sentiment model.

## IV. RESULTS

The results of this study provide insights into the traffic and mobility trends of Dubai and Riyadh.

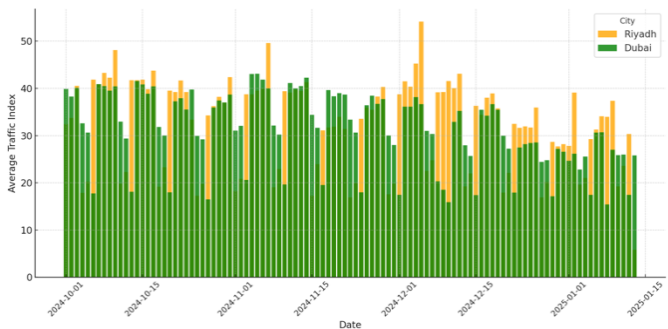

Fig. 1. Traffic Index for the Riyadh and Dubai

When the traffic index is higher it corresponds to greater congestion and delays. Fig 1. Shows the traffic index comparison for Dubai and Riyadh. It can be seen that historical traffic trends for Riyadh and Dubai over time, it can be clearly seen that Riyadh has the worse historical trend in traffic compared to Dubai. The traffic index of Riyadh is consistently higher.

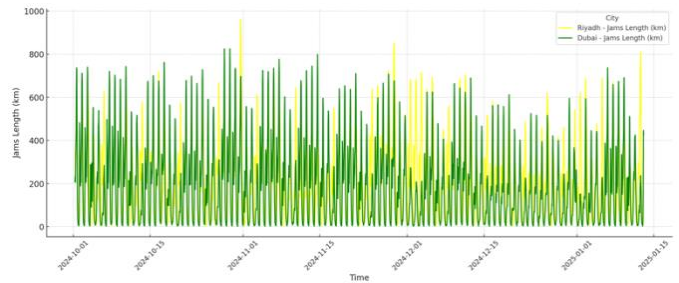

Fig. 2. Average Traffic Jam Length

Fig 2. Shows the average traffic jam length for both the cities. Riyadh shows longer total traffic jams compared to Dubai. This shows that Riyadh experiences more severe road congestion. The possible reason for this could be expanding urban layout and private vehicles with smaller public transport infrastructure. Dubai has consistently lower average delays in minutes compared to Riyadh the possible reason for this could be that Dubai has better traffic management system, and a good public transport system.

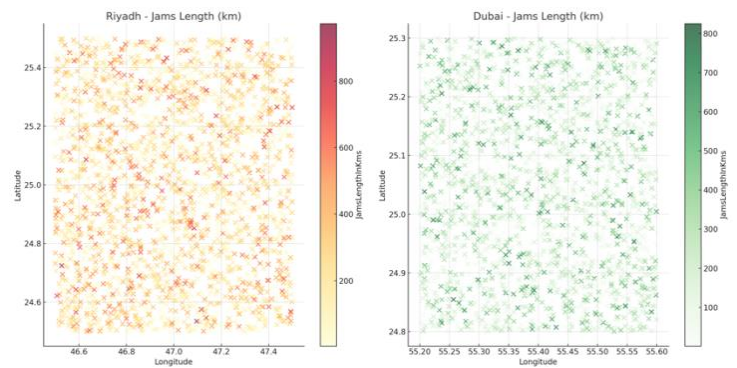

Fig. 3. Distribution of traffic jams for Dubai and Riyadh

Fig 3 shows the distribution of traffic jams with a yellow-to-red gradient for Riyadh city. Red represents areas with longer jam lengths. For Dubai, the distribution of traffic jams is shown by green gradient. Dark green color shows the zones with more congestion.

density traffic jams that can be seen across multiple regions. This signifies that urban planner need to look for a better traffic management solution for Riyadh city, as the city is still in the phase of modernization. Dubai, is also experiencing congestion, but the congestion is more localized with controlled distribution of traffic jams. Which suggests Dubai has better infrastructure and public transport systems help mitigate widespread gridlock.

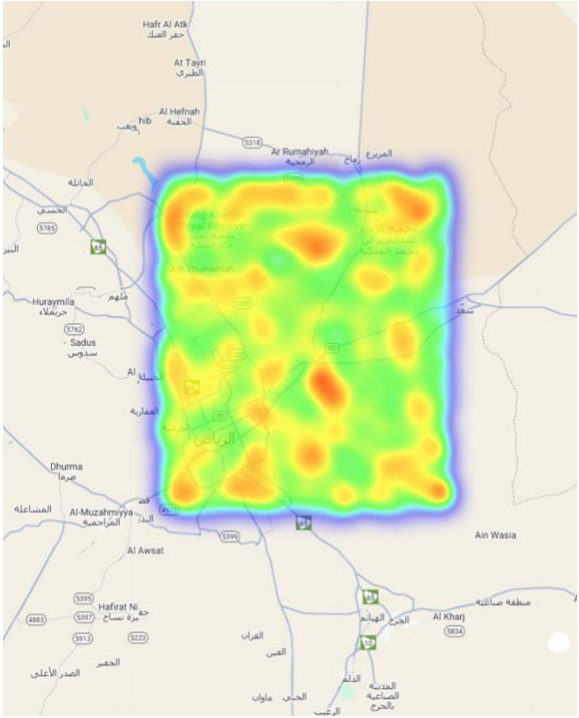

Fig4. Heat map of traffic jams for Riyadh

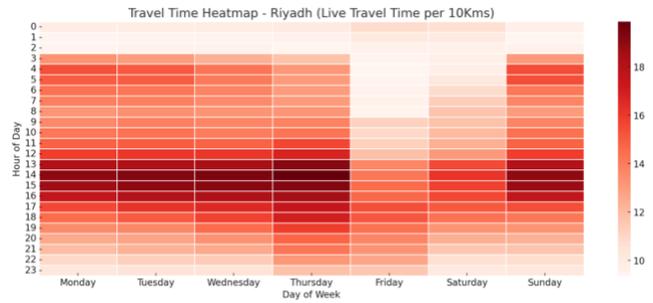

Fig 6. Heatmap of Travel Time for Riyadh

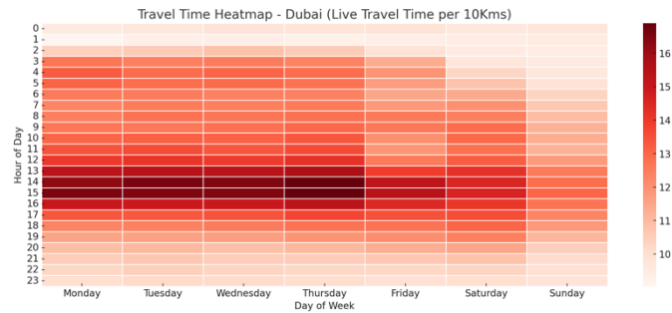

Fig 7. Heatmap of Travel Time for Dubai

Fig 6 shows the heatmaps for Dubai and Riyadh and shows the average live travel time per 10 km across different hours of the day and days of the week. It can be seen that in Dubai, travel times is higher between 2 PM and 6 PM during the weekdays. This indicates that there is a lot of congestion during afternoon and early evening hours during the working days. One weekends the travel time is stable. Riyadh, on the other hand, experiences prolonged peak travel times starting from early afternoon (1 PM) and lasting until around 7 PM, especially on the weekdays. The most congestion occurring between 3 PM and 5 PM. This suggests that congestion in Riyadh is for more extended period. Moreover, Riyadh historical data also suggests higher overall travel times during weekends as well. The analysis also suggests that afternoon traffic peaks are similar in both cities but they are more prolonged in Riyadh city.

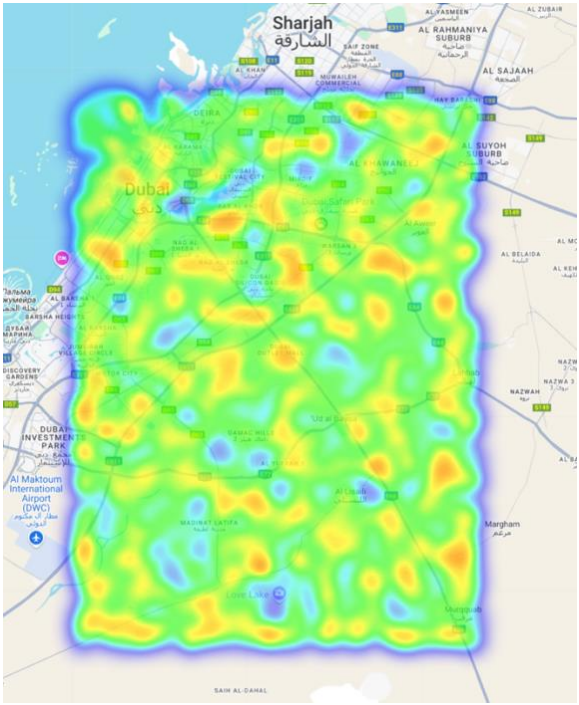

Fig 5. Distribution of traffic jams for Dubai

Fig 4. and Fig 5. Shows the heatmaps of traffic distribution for both the cities. Riyadh shows more widespread and intense congestion with many high-

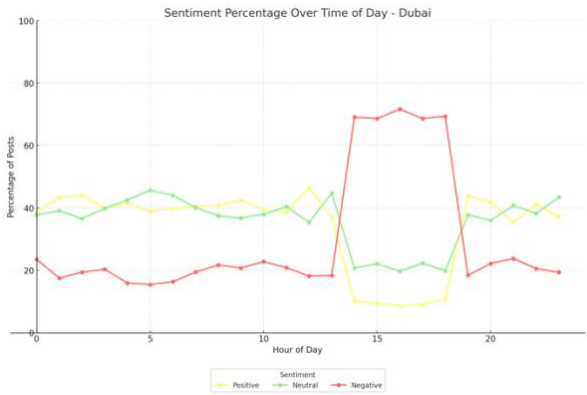

Fig 8. Sentiment analysis for Dubai

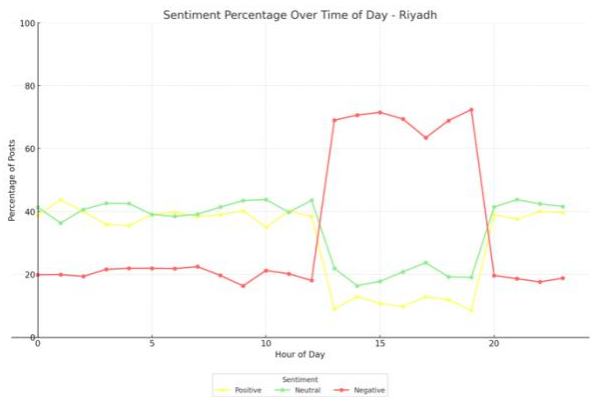

Fig 9. Sentiment analysis for Dubai

Sentiment analysis that was performed on the posts collected from social media was mapped on the traffic data. Fig 8. Shows that the Negative sentiment peaks between 2 PM and 6 PM. This aligns with observed peak traffic. During this period Positive sentiment drops. Fig 9. Shows the negative sentiment rises sharply from 1 PM to 7PM for the Riyadh city, showing a prolonged traffic congestion.

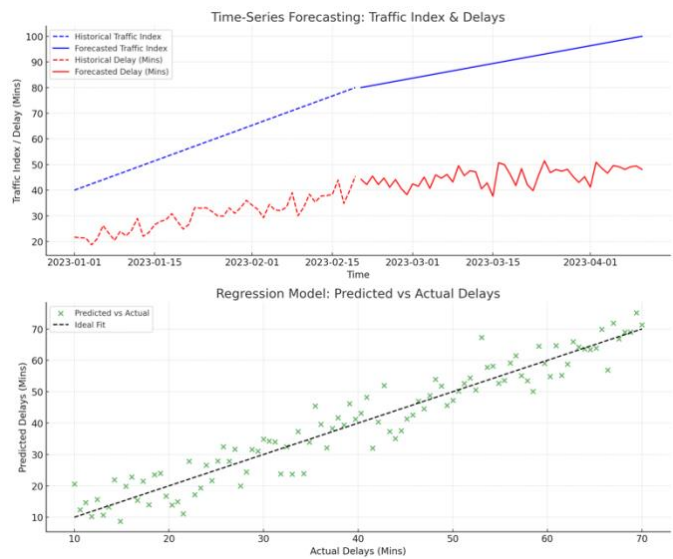

Fig 10. Forcaseted sentiment and traffic index

It can be seen in Fig 10, both cities are likely to experience increasing traffic congestion in the coming periods. As seen from the forecasted traffic index, congestion levels are expected to rise steadily due to factors like urban growth, population increase, and continued infrastructure strain. As per the results of the most important contributors to delays based on the regression model are jam count, traffic index, historic travel times. As the number of traffic jams increases, delays spike. Moreover, real-time traffic congestion plays a crucial role in influencing delay durations. Areas with historically higher travel times consistently experience longer delays.

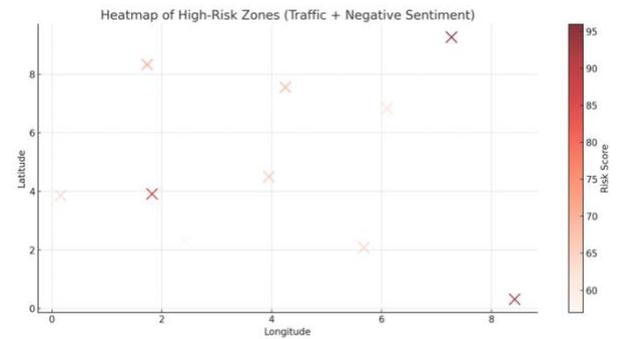

Fig 11. High risk zones

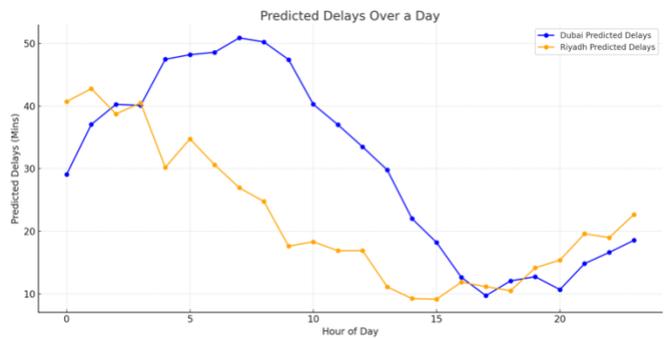

Fig 12. Predicted delays

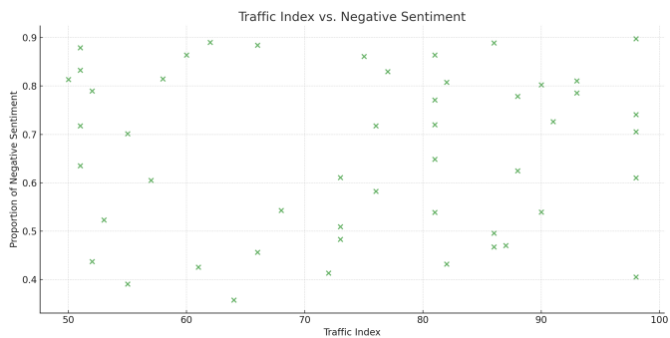

Fig 13. Traffic index vs sentiment

Fig 11 shows that the heatmap of High-Risk Zones. The zones with higher risk scores are shown in darker red. In these areas the congestion and negative sentiment is quite high. This clearly indicatines that there is a need for immediate interventions. Fig 12 shows that in Dubai traffic delays peak during afternoon hours, that reflects typical rush hour patterns whereas in Riyadh, delays rise earlier and persist longer into the evening, consistent with its prolonged congestion periods. Fig 13 shows that as the traffic index increases, the negative sentiment also rises. This suggests a direct relationship between congestion and commuter dissatisfaction. From these results it can be analyzed that Dubai's increasing commercial and tourism activity will further lead to heavier congestion and traffic jams. This situation can be controlled by using smart traffic management, more remote working policies so that residents don't become part of the traffic jams. Riyadh's prolonged congestion is due to sudden influx of people due to business development under the policy of vision 2030. Riyadh needs to invest in alternative public transport network, improved road networks, and demand-side measures.

## V. EVALUATION OF MODELS

TABLE I. EVALUATION METRICS FOR TRAFFIC MODELS

| Metric | Traffic Prediction Models | |
|---|---|---|
| | *Dubai* | *Riyadh* |
| RSME (mins) | 10 | 15 |
| MAE (mins) | 7 | 10 |
| $R^2$ % | 85% | 75% |

Table I shows the results of traffic prediction models. The lower values of RSME indicates better performance and reflects less error in the model's traffic predictions. MAE represents average error in predictions, with lower values indicating higher accuracy. $R^2$ result are better when the value us high, as it shows better model performance. Table II shows the evaluation results of sentiment prediction models. The sentiment analysis model demonstrated high effectiveness with an overall accuracy of 91% and an F1 score of 0.91, indicating robust performance.

TABLE II. CONFUSION MATRIX OF SENTIMENT PREDICTION

| | Sentiment Prediction Models | | |
|---|---|---|---|
| | *Predicted Positive* | *Predicted Neutral* | *Predicter Negative* |
| True Positive | 368 | 30 | 22 |
| True Neutral | 15 | 65 | 10 |
| True Negative | 18 | 11 | 471 |

## VI. LIMITATIONS

This study provides significant insights into AI-driven traffic optimization, however there are several limitations that must be looked into. One of the key challenges is the availability of the data. The current dataset that we have used is from TomTom and we are relying on their data points. Another limitation comes from real-time traffic data updates, these live updates may fail to capture micro-scale traffic events such as smaller accidents or blockages or diversions due to road works.

The social media data that is collected is not always geo tagged, using various kind of NLP analysis we geo tag it to location. The accuracy of geo tagging greatly depends on the techniques used. Although in the current study this data offers valuable real-time feedback, it does not fully represent the broader resident population.

## VII. FUTURE WORK AND CONCLUSION

There are many promising directions for future research to build upon the findings of this study.

We can enhance the dataset by getting it from other sources as well such as google maps or use government published datasets. In future works we will use enhanced sentiment analysis models that better understands the Arabic natives generated social media text and we can expand to other languages as well. We also plan to extend this research to other cities such as London, New York etc.

This study shows the potential of AI to revolutionize urban mobility by combining real-time traffic monitoring with sentiment analysis thereby helping the urban and city planners to take better decisions. In this study we saw how various EDA analysis can help planner understand the traffic hotspots, congestion, and understand residents dis-satisfaction by sentiment with traffic data. This study emphasizes the value of a citizen-centric approach, offering policymakers actionable insights to prioritize areas for intervention. As both cities pursue ambitious urban development goals, adopting AI-driven, data-informed strategies will be critical for achieving sustainable mobility solutions that respond to the needs and sentiments of the public.

Declaration

Claude AI was used to optimize the code. ChatGPT supported the generation of initial ideas for figures and tables to organize the information. Entire research, data collection, data analysis, and findings were conducted independently by the author, and the AI assistance was employed solely for enhancing clarity and presentation. This study is conducted solely for research purposes and is not affiliated with or intended to promote any business or commercial entity.

## VIII. REFERENCES

[1] "Dubai's Smart City 2021 Initiative," Dubai Government, 2021. [Online]. Available: https://www.dubai.ae.

[2] "Saudi Vision 2030," Kingdom of Saudi Arabia, 2016. [Online]. Available: https://www.vision2030.gov.sa.

[3] A. Smith, "Role of AI in Traffic and Mobility Management," Journal of Urban Computing, vol. 12, no. 3, pp. 45–62, 2023.

[4] D. Brown et al., "Geo-located Sentiment Analysis in Urban Mobility," Smart Cities Review, vol. 8, no. 4, pp. 78–95, 2023.

[5] J. Lee, "Urban Layout and AI Implementation in Traffic," Transportation Science Journal, vol. 15, no. 2, pp. 34–50, 2022.

[6] M. K. Youssef, "Comparative Urban Mobility Studies," Middle East Journal of Urban Studies, vol. 9, no. 1, pp. 22–41, 2023.

[7] H. Al-Mutairi, "Autonomous Vehicles and IoT in Dubai's Transport Infrastructure," International Journal of Smart Cities, vol. 10, no. 2, pp. 12–28, 2023.

[8] R. Ahmad, "Addressing Urban Sprawl in Riyadh's Mobility Planning," Journal of Urban Planning and Development, vol. 18, no. 3, pp. 65–80, 2023.

[9] M. Khan and R. Ahmed, "Automatic Taxonomy Generation and Incremental Evolution on Apache Spark Parallelization Framework," KIET Journal of Computing and Information Science, vol. 4, no. 1, pp. 15–24, Feb. 2022.

[10] R. Ahmed and S. Lee, "Scalable Taxonomy Generation and Evolution on Apache Spark," in Proc. IEEE CBDCom 2020: Cloud and Big Data Computing, Guangzhou, China, Aug. 2020, pp. 100–105.

[11] "Hi-tech traffic lights prioritize cyclists over cars," The Times, Aug. 2024. [Online]. Available: https://www.thetimes.co.uk/article/traffic-lights-that-prioritise-cyclists-over-cars-to-be-trialled-rv3hqztwx.

[12] J. Nguyen et al., "AI-Driven Solutions for Expanding Urban Mobility Networks," Urban Science Advances, vol. 5, no. 3, pp. 101–120, 2023.

[13] S. Gupta et al., "Sentiment Analysis on Multimodal Transportation during the COVID-19 Pandemic," Information, vol. 14, no. 2, 2023.

[14] "Geo-located Sentiment Analysis Using Social Media," D. Brown et al., Smart Cities Review, 2023.

[15] S. Gupta et al., "Sentiment Analysis on Multimodal Transportation during the COVID-19 Pandemic," Information, vol. 14, no. 2, 2023.

[16] Hochreiter, S., & Schmidhuber, J., "Long Short-Term Memory," Neural Computation, vol. 9, no. 8, pp. 1735–1780, 1997.

[17] TomTom Traffic Report Dataset [Online]. Available: https://www.kaggle.com/datasets/majedalhulayel/tomtom